# ON ENHANCING THE PERFORMANCE OF NEAREST NEIGHBOUR CLASSIFIERS USING HASSANAT DISTANCE METRIC

Mouhammd Alkasassbeh[1], Ghada A. Altarawneh[2], Ahmad B. Hassanat[3]
[1,3]IT Department, [2]Accounting Department, Mutah University, Mutah – Karak, Jordan, 61710
[1]malkasasbeh@gmail.com, [3]ahmad.hassanat@gmail.com

## ABSTRACT

We showed in this work how the Hassanat distance metric enhances the performance of the nearest neighbour classifiers. The results demonstrate the superiority of this distance metric over the traditional and most-used distances, such as Manhattan distance and Euclidian distance. Moreover, we proved that the Hassanat distance metric is invariant to data scale, noise and outliers. Throughout this work, it is clearly notable that both ENN and IINC performed very well with the distance investigated, as their accuracy increased significantly by 3.3% and 3.1% respectively, with no significant advantage of the ENN over the IINC in terms of accuracy. Correspondingly, it can be noted from our results that there is no optimal algorithm that can solve all real-life problems perfectly; this is supported by the no-free-lunch theorem.

## INTRODUCTION

The nearest neighbour (KNN) classifier is one of the most used and well-known approaches for performing recognition tasks since it was first introduced in 1951 by (Fix & Hodges, 1951) and later developed by (Cover & Hart, 1967). This approach is one of the simplest and oldest methods used in data mining (DM) and pattern classification. Moreover, it is considered one of the top 10 methods in DM (Wu, 2010). It often yields efficient performance and, in certain cases, its accuracy is greater than state-of the-art classifiers (Hamamoto et al., 1997)(Alpaydin, 1997).The KNN classifier categorizes an unlabelled test example using the label of the majority of examples among its k-nearest (most similar) neighbours in the training set. The similarity depends on a specific distance metric, therefore, the performance of the classifier depends significantly on the distance metric used (Weinberger & Saul, 2009).

A large number of similarity measures are proposed in the literature, perhaps the most famous and well known being the Euclidean distance (ED) stated by Euclid two thousand years ago. The ED has been receiving increased attention, because many applications, such as bioinformatics, dimensionality reduction in machine learning, statistics, and many others have all become very active research areas, which are mainly dependent on ED (Nathan Krislock, 2012).

In addition, over the last century, great efforts have been made to find new metrics and similarity measures to satisfy the needs of different applications. New similarity measures are needed, in particular, for use in distance learning (Yang, 2006), where classifiers such as the k-nearest neighbour (KNN) are heavily dependent upon choosing the best distance. Optimizing the distance metric is valuable in several computer vision tasks, such as object detection, content-based image retrieval, image segmentation and classification.

The similarity measures that are used the most are Euclidean (ED) and Manhattan distances (MD); both assume the same weight to all directions. In addition, the difference between vectors at each dimension might approach infinity to imply dissimilarity (Bharkad & Kokare, 2011). Therefore, such types of distances are heavily affected by the different scale of the data, noise and outliers. To solve those problems, Hassanat proposed an interesting distance metric (Hassanat, 2014), which is invariant to the different scales in multi dimensions data.

The main purpose of this work is to equip some of the nearest neighbour classifiers with Hassanat distance, attempting to enhance their performance. The rest of this paper describes some of the nearest neighbour classifiers to be enhanced, in addition to describing the Hassanat metric. The third section describes the data set used for experiments and discusses the results, focusing on applying the new metric which is used by some nearest neighbour classifiers.

## NEAREST NEIGHBOUR CLASSIFIERS

We will describe the traditional KNN, Inverted Indexes of Neighbours Classifier (IINC) (Jirina & Jirina, 2008; Jirina & Jirina, 2011) and Ensemble Nearest Neighbour classifiers (ENN) (Hassanat, 2014).

### KNN

KNN is a very simple yet effective classifier that is used for pattern classification. It categorizes an unlabelled test example using the label of the majority of examples among its k-nearest (most similar) neighbours in the training set (see Figure 1). The similarity depends on a specific distance metric, normally ED or MD, therefore, the performance of the classifier depends significantly on the distance metric used. Because of its simplicity and popularity it has been used extensively in pattern recognition, machine learning, text categorization, data mining, object recognition, etc. (Kataria & Singh, 2013)(Bhatia & Vandana, 2010) and (Hassanat, 2009). However, it has some limitations, such as memory requirement and time complexity, because it is fully dependent on every example in the training set and choosing the optimal k neighbours in advance.

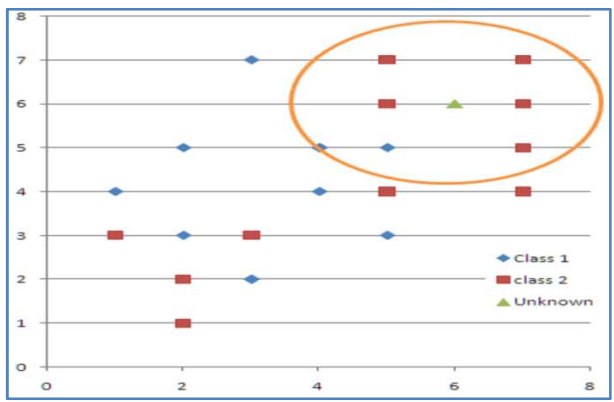

Fig. 1. KNN classifies the unknown example to class 2 based on the k=7 neighbours

**Inverted Indexes of Neighbours Classifier (IINC).**

The inverted indexes of neighbours classifier IINC (Jirina & Jirina, 2008), (Jirina & Jirina, 2010)and (Jirina & Jirina, 2011) is one of the best attempts found in the literature to solve the optimal k problems related to the KNN classifier. The aim of their work was to increase the accuracy of the classifier by using all the neighbours in the training set, so k has become the whole number of examples in the training set, rewarding the nearest neighbours by adding some heavyweight, and penalizing the furthest one by giving them lightweight. Moreover, the first nearest neighbour of the point, for example x, has the biggest influence on what class point x goes to. The IINC approach is mainly based on the hypothesis that the influence, the weight of a neighbour, is proportional to the distance from the query point.

The IINC algorithm works as follows: the distances between the test point and the other points in the training set are calculated, and then sorted in ascending order. The summation of the inverted indexes is then calculated for each class using Eq (1). The probability of each class is then calculated using Eq (2). Obviously, the class with the highest probability is then predicted.

$$S_c = \sum_{i=1(c)}^{L_c} \frac{1}{i} \quad (1)$$

Where $L_c$ is the number of points of class *c*, *i* is the order of the point in the training set after sorting the distances.

The probability of a test point x belonging to a class c is estimated by:

$$P(x|c) = \frac{S_c}{S} \quad (2)$$

where $S = \sum_{i=1}^{N} \frac{1}{i}$ and N is the number of examples in the training set.

To visualize the IINC, cover all the points in Figure 1 by a large circle that fits all the points.

**ENN**

The ENN classifier (Hassanat, 2014) uses an ensemble learning approach based on the same nearest neighbour rule. Fundamentally, the traditional KNN classifier is used each time with a different K. Starting from k=1 to k= the square root of the number of examples in the training set, each classifier votes for a specific class. Then it uses the weighted sum rule to identify the class, i.e. the class with the highest sum (by 1-NN, 3-NN, 5-NN… √n-NN) is chosen.

ENN uses √n down to k=1 with only odd numbers to increase the speed of the algorithm by avoiding the even classifiers and to avoid the chance of two different classes having the same number of votes. The used weight is expressed by:

$$w(k) = \frac{1}{\log_2(1+k)} \quad (3)$$

When a test point is compared with all examples, using some distance function, an array (A) is created to hold the nearest √n classes, and the weighted sum (WS) rule is defined for each class using:

$$WS_c = \sum_{k=1}^{\sqrt{n}} \sum_{i=1}^{k} \begin{cases} w(i), & A_i = c \\ 0, & otherwise \end{cases}, k = k+2 \quad (4)$$

where, for each class, the outer sum represents the KNN classifier for each odd k, and the inner sum calculates the weights for each classifier.

The predicted class is the one with the maximum weighted sum:

$$class = \underset{c}{\operatorname{argmax}} WS_c \quad (5)$$

To visualize the ENN classifier, assume that there are 25 points in a 2D feature space belonging to 2 different classes (0 and 1), and one unknown point (the green triangle) as shown in Figure 1. The ensemble system uses the 1-NN, 3-NN and 5-NN classifiers altogether using the WS rule to find the class of the "*green triangle*", which in this example and according to the ENN is predicted to be class 1 "*red square*".

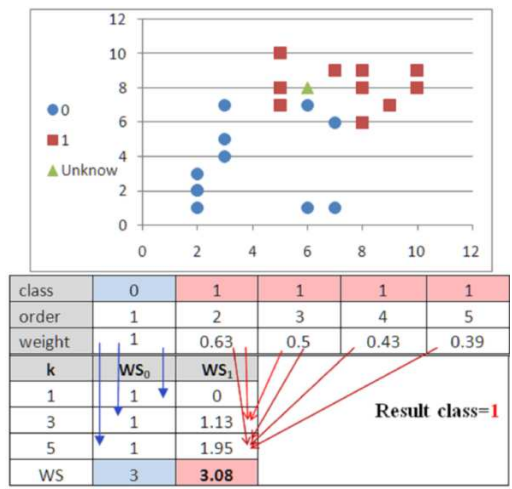

Fig. 2. Simple example showing the ENN classifier (Hassanat, 2014)

**Hassanat distance metric**

The similarity function between any two points using Hassanat Metric (Hassanat, 2014) is written as:

$$D(A_i, B_i) = \begin{cases} 1 - \frac{1+min(A_i,B_i)}{1+max(A_i,B_i)}, & min(A_i,B_i) \geq 0 \\ 1 - \frac{1+min(A_i,B_i)+|min(A_i,B_i)|}{1+max(A_i,B_i)+|min(A_i,B_i)|}, & min(A_i,B_i) < 0 \end{cases} \quad (6)$$

Along the vectors dimensions the distance is:

$$D_{Hassanat}(A, B) = \sum_{i=1}^{m}\bigl(D(A_i, B_i)\bigr) \quad (7)$$

where A and B are both vectors with size m. $A_i$ and $B_i$ are real numbers.

This metric is invariant to similarity measure is invariant to different scale, noise and outliers because it is bounded by the interval [0, 1[. It reaches to 1 only when the maximum value approaches infinity, or when the minimum value approaches minus infinity. This is shown by Figure 3 and equation 8. This means that the more two values are similar, the nearest to zero the distance will be, and the more they are dissimilar, the nearest to one the distance will be.

$$\lim_{max(A_i,B_i) \to \infty}\bigl(D(A_i, B_i)\bigr) = \lim_{min(A_i,B_i) \to -\infty}\bigl(D(A_i, B_i)\bigr) = 1 \quad (8)$$

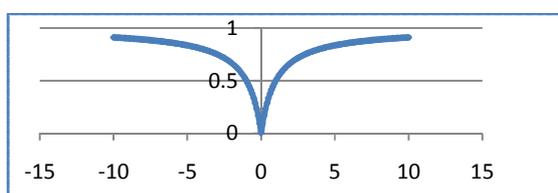

Fig. 3. Representation of Hassanat distance metric between the points 0 and n, where n belongs to [-10, 10](Hassanat, 2014)

**DATA USED FOR OUR EXPERIMENTS**

For evaluation of the efficiency of the Hassanat distance when used with some classifiers, twenty eight different data sets were chosen to represent real life classification problems, taken from the UCI Machine Learning Repository(Bache & Lichman, 2013). This is a collection of databases, domain theories, and data generators that are used by the machine learning community. Since the database was created in 1987 by David Aha and fellow graduate students at UC Irvine, it has been widely used by researchers, students and educators all over the world as a primary source of machine learning data sets. Readers can know more about each dataset using the following link: http://mlr.cs.umass.edu/ml/. Table 1 depicts the data sets we used in our work.

Table 1. Description of the real world data sets used.

| Name | #E | #F | #C | data type | Min | Max |
|---|---|---|---|---|---|---|
| Heart | 270 | 25 | 2 | pos integer | 0 | 564 |
| Balance | 625 | 4 | 3 | pos integer | 1 | 5 |
| Cancer | 683 | 9 | 2 | pos integer | 0 | 9 |
| German | 1000 | 24 | 2 | pos integer | 0 | 184 |
| Liver | 345 | 6 | 2 | pos integer | 0 | 297 |
| Vehicle | 846 | 18 | 4 | pos integer | 0 | 1018 |
| Vote | 399 | 10 | 2 | pos integer | 0 | 2 |
| BCW | 699 | 10 | 2 | pos integer | 1 | 13454352 |
| Haberman | 306 | 3 | 2 | pos integer | 0 | 83 |
| Letter rec. | 20000 | 16 | 26 | pos integer | 0 | 15 |
| Wholesale | 440 | 7 | 2 | pos integer | 1 | 112151 |
| Australian | 690 | 42 | 2 | pos real | 0 | 100001 |
| Glass | 214 | 9 | 6 | pos real | 0 | 75.41 |
| Sonar | 208 | 60 | 2 | pos real | 0 | 1 |
| Wine | 178 | 13 | 3 | pos real | 0.13 | 1680 |
| EEG | 14980 | 14 | 2 | pos real | 86.67 | 715897 |
| Parkinson | 1040 | 27 | 2 | pos real | 0 | 1490 |
| Iris | 150 | 4 | 3 | pos real | 0.1 | 7.9 |
| Diabetes | 768 | 8 | 2 | real & integer | 0 | 846 |
| Monkey1 | 556 | 17 | 2 | binary | 0 | 1 |
| Ionosphere | 351 | 34 | 2 | real | -1 | 1 |
| Phoneme | 5404 | 5 | 2 | real | -1.82 | 4.38 |
| Segmen | 2310 | 19 | 7 | real | -50 | 1386.33 |
| Vowel | 528 | 10 | 11 | real | -5.21 | 5.07 |
| Wave21 | 5000 | 21 | 3 | real | -4.2 | 9.06 |
| Wave40 | 5000 | 40 | 3 | real | -3.97 | 8.82 |
| Banknote | 1372 | 4 | 2 | real | -13.77 | 17.93 |
| QSAR | 1055 | 41 | 2 | real | -5.256 | 147 |

Where #E means number of examples, and #F means number of features and #C means number of classes.

**RESULTS AND DISCUSSION**

Each data set is divided into two data sets, one for training, and the other for testing. 30% of the data set is used for testing, and 70% of the data is for training. Each classifier is used to classify the test samples using Manhattan distance (see Table 2). All experiments were repeated using Hassanat distance (see Table 3). 30% of the data, which were used as a test sample, were chosen randomly and each experiment on each data set was repeated 10 times to obtain random examples for testing and training. Tables 2 and 3 show the results of the experiments. The accuracy of each classifier in each data set is the average of 10 rounds.

Table 2. Accuracy of different nearest neighbour classifiers using Manhattan distance without data normalization

| Data set | 1NN | 3NN | 5NN | 7NN | 9NN | √nNN | IINC | ENN |
|---|---|---|---|---|---|---|---|---|
| Australian | 0.69 | 0.71 | 0.71 | 0.71 | 0.71 | 0.69 | 0.71 | **0.72** |
| Balance | 0.79 | 0.81 | 0.83 | 0.85 | 0.86 | **0.88** | 0.87 | 0.87 |
| Banknote | **1.00** | **1.00** | **1.00** | **1.00** | **1.00** | 0.99 | **1.00** | 1.00 |
| BCW | 0.61 | 0.61 | 0.61 | 0.62 | 0.63 | 0.66 | **0.66** | 0.64 |
| Cancer | 0.96 | **0.97** | 0.96 | 0.97 | 0.97 | 0.96 | 0.95 | 0.97 |
| Diabetes | 0.70 | 0.71 | 0.73 | 0.73 | 0.73 | **0.75** | 0.73 | 0.74 |
| EEG | **0.97** | 0.97 | 0.96 | 0.95 | 0.95 | 0.86 | 0.95 | 0.93 |

| German | 0.68 | 0.72 | 0.72 | 0.73 | 0.73 | 0.73 | 0.73 | **0.74** |
|---|---|---|---|---|---|---|---|---|
| Glass | **0.71** | 0.69 | 0.68 | 0.66 | 0.66 | 0.65 | 0.69 | 0.69 |
| Haberman | 0.67 | 0.70 | 0.73 | 0.73 | **0.76** | **0.76** | 0.75 | 0.72 |
| Heart | 0.64 | 0.70 | 0.70 | 0.69 | **0.70** | 0.69 | 0.69 | 0.69 |
| Ionosphere | **0.90** | 0.89 | 0.89 | 0.89 | 0.87 | 0.85 | 0.85 | 0.89 |
| Iris | 0.95 | **0.96** | 0.95 | 0.95 | 0.96 | 0.95 | 0.96 | 0.96 |
| Letter rec. | 0.95 | 0.95 | 0.95 | 0.95 | 0.94 | 0.83 | **0.95** | 0.94 |
| Liver | 0.60 | 0.62 | 0.65 | **0.68** | 0.67 | 0.67 | 0.66 | 0.66 |
| Monkey1 | 0.79 | 0.84 | 0.91 | 0.95 | **0.96** | 0.92 | 0.92 | 0.94 |
| Parkinson | 0.82 | 0.82 | 0.83 | 0.82 | 0.82 | 0.80 | **0.84** | 0.84 |
| Phoneme | **0.89** | 0.88 | 0.87 | 0.87 | 0.86 | 0.83 | 0.87 | 0.87 |
| QSAR | 0.81 | 0.82 | 0.82 | 0.83 | 0.82 | 0.79 | **0.84** | 0.83 |
| Segmen | **0.97** | 0.96 | 0.96 | 0.95 | 0.94 | 0.90 | 0.95 | 0.95 |
| Sonar | **0.83** | 0.80 | 0.78 | 0.73 | 0.70 | 0.68 | 0.83 | 0.82 |
| Vehicle | 0.67 | 0.68 | 0.68 | 0.67 | 0.66 | 0.65 | 0.68 | 0.69 |
| Vote | 0.91 | 0.93 | 0.93 | **0.94** | **0.94** | 0.93 | 0.93 | 0.93 |
| Vowel | **0.98** | 0.94 | 0.87 | 0.78 | 0.71 | 0.56 | 0.96 | 0.93 |
| Waveform21 | 0.78 | 0.80 | 0.82 | 0.83 | 0.83 | 0.85 | 0.84 | **0.85** |
| Waveform40 | 0.75 | 0.80 | 0.80 | 0.82 | 0.83 | **0.86** | 0.84 | 0.85 |
| Wholesale | 0.88 | 0.89 | 0.90 | 0.91 | 0.91 | **0.91** | 0.91 | 0.91 |
| Wine | **0.81** | 0.74 | 0.73 | 0.74 | 0.75 | 0.74 | 0.80 | 0.79 |
| **Average** | 0.81 | 0.82 | 0.82 | 0.82 | 0.82 | 0.80 | **0.83** | **0.83** |

As can be seen from Table 2, the results in general are worse than those in (Hassanat, 2014), because we repeated the experiments without normalizing the data sets using Manhattan (same) distance. Except for some datasets such as the "EEG", whose accuracy increased significantly from 84% to 97%, this result proves that data might be harmed by normalization, and there is a need for a distance metric that is not affected by the data scale, and therefore does not need data normalization. These results confirm some results in (Hassanat, 2014), such as the superiority of the IINC and the ENN classifiers, in terms of being independent from choosing the optimal k neighbours. On the other hand, after employing Hassanat distance with the nearest neighbour classifiers on the same data sets and without normalization, we obtained the results depicted in Table 3.

Table 3: Accuracy of nearest neighbour classifiers using Hassanat distance without data normalization

| Dataset | 1NN | 3NN | 5NN | 7NN | 9NN | √nNN | IINC | ENN |
|---|---|---|---|---|---|---|---|---|
| Australian | 0.82 | 0.86 | 0.87 | **0.87** | 0.86 | 0.86 | 0.86 | 0.87 |
| Balance | 0.82 | 0.82 | 0.83 | 0.85 | 0.86 | 0.88 | **0.89** | 0.86 |
| Banknote | **1.00** | **1.00** | **1.00** | **1.00** | 1.00 | 0.99 | **1.00** | **1.00** |
| BCW | 0.96 | **0.97** | 0.97 | 0.97 | 0.96 | 0.96 | 0.96 | 0.97 |
| Cancer | 0.97 | 0.97 | 0.97 | 0.97 | 0.97 | 0.97 | 0.97 | **0.97** |
| Diabetes | 0.67 | 0.70 | 0.71 | 0.73 | 0.73 | **0.75** | 0.72 | 0.73 |
| EEG | **0.97** | 0.97 | 0.96 | 0.95 | 0.95 | 0.86 | 0.94 | 0.93 |
| German | 0.68 | 0.72 | 0.73 | 0.73 | 0.74 | 0.73 | 0.72 | **0.75** |
| Glass | 0.68 | 0.68 | 0.65 | 0.66 | 0.66 | 0.66 | 0.69 | **0.70** |
| Haberman | 0.65 | 0.70 | 0.71 | 0.71 | **0.72** | 0.72 | 0.72 | 0.71 |
| Heart | 0.79 | 0.80 | 0.82 | **0.83** | 0.82 | 0.82 | 0.82 | 0.82 |
| Ionosphere | **0.91** | 0.91 | 0.90 | 0.90 | 0.91 | 0.89 | 0.89 | 0.90 |
| Iris | 0.94 | 0.95 | 0.95 | **0.96** | 0.96 | 0.95 | 0.95 | 0.95 |
| Letter rec. | 0.95 | 0.94 | 0.94 | 0.94 | 0.94 | 0.82 | **0.95** | 0.94 |
| Liver | 0.61 | 0.65 | 0.67 | 0.66 | 0.66 | **0.68** | 0.67 | 0.67 |
| Monkey1 | 0.78 | 0.84 | 0.91 | 0.94 | 0.96 | 0.91 | 0.92 | 0.94 |
| Parkinson | 0.92 | 0.95 | 0.96 | 0.96 | 0.96 | 0.97 | **0.98** | 0.98 |
| Phoneme | 0.90 | 0.89 | 0.87 | 0.87 | 0.87 | 0.84 | 0.87 | **0.87** |
| QSAR | 0.82 | 0.84 | 0.84 | 0.85 | 0.84 | 0.83 | **0.85** | 0.85 |
| Segmen | 0.96 | 0.95 | 0.95 | 0.95 | 0.95 | 0.90 | 0.95 | 0.95 |
| Sonar | 0.86 | 0.84 | 0.83 | 0.79 | 0.75 | 0.73 | 0.85 | **0.86** |
| Vehicle | 0.66 | 0.67 | 0.67 | 0.67 | 0.67 | 0.67 | 0.68 | **0.69** |
| Vote | 0.92 | 0.92 | 0.92 | 0.92 | 0.92 | 0.91 | 0.92 | **0.93** |
| Vowel | **0.97** | 0.92 | 0.85 | 0.76 | 0.69 | 0.54 | 0.94 | 0.94 |
| Waveform21 | 0.75 | 0.79 | 0.81 | 0.82 | 0.82 | **0.85** | 0.83 | 0.84 |
| Waveform40 | 0.72 | 0.77 | 0.79 | 0.80 | 0.81 | **0.84** | 0.83 | 0.84 |
| wholesale | 0.87 | 0.88 | 0.89 | 0.90 | **0.90** | 0.90 | 0.90 | 0.89 |
| Wine | **0.97** | 0.97 | 0.96 | 0.96 | 0.96 | 0.96 | 0.97 | 0.97 |
| **Average** | 0.84 | 0.85 | 0.86 | 0.85 | 0.85 | 0.84 | **0.87** | **0.87** |

By comparing columns in Tables 2 and 3, the significant increase in the performance of each algorithm can be observed. Table 4 illustrates the increase in the accuracy of each algorithm after applying Hassanat distance; this enhancement proves that Hassanat distance is not affected by the scale of the data.

Table 4. Results of each algorithm after applying Hassanat distance

| Dataset | 1NN | 3NN | 5NN | 7NN | 9NN | √nNN | IINC | ENN |
|---|---|---|---|---|---|---|---|---|
| Australian | 0.13 | 0.15 | 0.16 | 0.16 | 0.16 | 0.16 | 0.15 | 0.14 |
| Balance | 0.04 | 0.01 | 0.00 | 0.00 | 0.00 | 0.00 | 0.01 | -0.01 |
| Banknote | 0.00 | 0.00 | 0.00 | 0.00 | 0.00 | 0.00 | 0.00 | 0.00 |
| BCW | 0.35 | 0.36 | 0.36 | 0.34 | 0.33 | 0.30 | 0.30 | 0.32 |
| Cancer | 0.01 | 0.00 | 0.00 | 0.00 | 0.01 | 0.01 | 0.02 | 0.01 |
| Diabetes | -0.02 | -0.01 | -0.02 | 0.00 | -0.01 | 0.00 | -0.01 | -0.01 |
| EEG | 0.00 | 0.00 | 0.00 | 0.00 | 0.00 | 0.00 | 0.00 | 0.00 |
| German | 0.00 | 0.00 | 0.01 | 0.01 | 0.01 | 0.00 | -0.01 | 0.00 |
| Glass | -0.03 | -0.01 | -0.03 | 0.00 | 0.00 | 0.01 | -0.01 | 0.01 |
| Haberman | -0.02 | -0.01 | -0.02 | -0.02 | -0.04 | -0.04 | -0.03 | -0.01 |
| Heart | 0.15 | 0.10 | 0.12 | 0.14 | 0.11 | 0.13 | 0.13 | 0.13 |
| Ionosphere | 0.01 | 0.01 | 0.01 | 0.02 | 0.03 | 0.04 | 0.04 | 0.02 |
| Iris | -0.01 | -0.01 | 0.00 | 0.01 | 0.00 | 0.00 | -0.01 | -0.01 |
| Letter rec. | -0.01 | -0.01 | -0.01 | -0.01 | -0.01 | 0.00 | 0.00 | 0.00 |
| Liver | 0.01 | 0.03 | 0.02 | -0.01 | -0.01 | 0.02 | 0.02 | 0.01 |
| Monkey1 | -0.01 | 0.00 | 0.00 | 0.00 | 0.00 | -0.02 | -0.01 | 0.00 |
| Parkinson | 0.11 | 0.13 | 0.13 | 0.15 | 0.14 | 0.17 | 0.14 | 0.14 |
| Phoneme | 0.00 | 0.01 | 0.00 | 0.00 | 0.00 | 0.00 | 0.00 | 0.00 |
| QSAR | 0.01 | 0.02 | 0.02 | 0.02 | 0.02 | 0.04 | 0.01 | 0.02 |
| Segmen | -0.01 | -0.01 | -0.01 | 0.00 | 0.01 | -0.01 | 0.00 | 0.00 |
| Sonar | 0.02 | 0.04 | 0.05 | 0.06 | 0.05 | 0.05 | 0.02 | 0.04 |
| Vehicle | -0.01 | -0.01 | 0.00 | 0.01 | 0.01 | 0.01 | 0.01 | 0.00 |
| Vote | 0.01 | 0.00 | -0.01 | -0.01 | -0.01 | -0.02 | -0.01 | -0.01 |
| Vowel | -0.01 | -0.01 | -0.03 | -0.03 | -0.02 | -0.01 | -0.01 | 0.00 |
| Waveform21 | -0.03 | -0.01 | -0.01 | -0.01 | -0.01 | 0.00 | -0.01 | -0.01 |
| Waveform40 | -0.03 | -0.03 | -0.02 | -0.02 | -0.02 | -0.02 | -0.01 | -0.01 |
| Wholesale | -0.01 | -0.02 | -0.01 | -0.01 | -0.01 | -0.01 | -0.01 | -0.02 |
| Wine | 0.16 | 0.23 | 0.23 | 0.22 | 0.22 | 0.22 | 0.17 | 0.18 |
| **Average** | **0.03** | **0.03** | **0.04** | **0.04** | **0.04** | **0.04** | **0.03** | **0.03** |

The average row in Table 4 confirms that using Hassanat distance enhances the performance of the nearest neighbour algorithms by 2.9% to 3.8%. The same table also shows a significant improvement in most algorithms in most data sets, such as the 30% to 35.9% boost in the accuracy of the data set BCW. Although the increase in performance is the theme, sometimes when the performance is decreased, however, these degrades were not beyond 4.1%.

It can be noted from Table 3 that both ENN and IINC performed very well with Hassanat distance, as their accuracy increased significantly by 3.3% and 3.1% respectively, with no significant advantage of the ENN over the IINC in terms of accuracy, and this result is confirmed by (Hassanat, 2014). Also, it can be noted from Tables 2,3 and 4 that there is no optimal algorithm that can solve all real life problems perfectly; this conclusion is supported by the no-free-lunch theorem (Duda et al., 2001). However, some algorithms, such as ENN and IINC using Hassanat distance, give more stable results than the others, i.e. if they are not the best to classify a specific (problem) data set, they are not the worst, and their results are very close to the best-gained results. This type of stability is illustrated by Table 5, where we record the absolute difference between each result and the best result within a specific data set.

Table 5. The performance of each algorithm with respect to the best result for each data set

| Dataset | Best result | 1NN | 3NN | 5NN | 7NN | 9NN | √nNN | IINC | ENN |
|---|---|---|---|---|---|---|---|---|---|
| Australian | 0.87 | 0.05 | 0.01 | 0 | 0 | 0.01 | 0.01 | 0.01 | 0 |
| Balance | 0.89 | 0.06 | 0.06 | 0.06 | 0.04 | 0.02 | 0 | 0 | 0.02 |
| Banknote | 1 | 0 | 0 | 0 | 0 | 0 | 0.01 | 0 | 0 |
| BCW | 0.97 | 0.01 | 0 | 0 | 0 | 0.01 | 0 | 0 | 0 |
| Cancer | 0.97 | 0.01 | 0.01 | 0.01 | 0 | 0 | 0 | 0 | 0 |
| Diabetes | 0.75 | 0.08 | 0.05 | 0.04 | 0.02 | 0.02 | 0 | 0.03 | 0.02 |
| EEG | 0.97 | 0 | 0 | 0.01 | 0.02 | 0.02 | 0.11 | 0.03 | 0.04 |
| German | 0.75 | 0.07 | 0.02 | 0.01 | 0.01 | 0.01 | 0.02 | 0.03 | 0 |
| Glass | 0.7 | 0.02 | 0.01 | 0.04 | 0.04 | 0.04 | 0.04 | 0.01 | 0 |
| Haberman | 0.72 | 0.07 | 0.02 | 0.01 | 0.01 | 0 | 0.01 | 0 | 0.01 |
| Heart | 0.83 | 0.04 | 0.03 | 0.01 | 0 | 0.01 | 0.01 | 0.01 | 0.02 |
| Ionosphere | 0.91 | 0 | 0 | 0.01 | 0.01 | 0 | 0.02 | 0.02 | 0.01 |
| Iris | 0.96 | 0.02 | 0.01 | 0.01 | 0 | 0 | 0.01 | 0.01 | 0.01 |
| Letter-rec. | 0.95 | 0 | 0.01 | 0.01 | 0.01 | 0.01 | 0.13 | 0 | 0.01 |
| Liver | 0.68 | 0.07 | 0.03 | 0.01 | 0.02 | 0.02 | 0 | 0.01 | 0.01 |
| Monkey1 | 0.96 | 0.18 | 0.12 | 0.05 | 0.02 | 0 | 0.06 | 0.05 | 0.02 |
| Parkinson | 0.98 | 0.06 | 0.03 | 0.02 | 0.02 | 0.02 | 0.01 | 0 | 0.01 |
| Phoneme | 0.9 | 0 | 0.01 | 0.02 | 0.03 | 0.03 | 0.06 | 0.03 | 0.03 |
| QSAR | 0.85 | 0.03 | 0.01 | 0.01 | 0.01 | 0.01 | 0.02 | 0 | 0 |
| Segmen | 0.96 | 0 | 0.01 | 0.01 | 0.01 | 0.01 | 0.06 | 0.01 | 0.01 |
| Sonar | 0.86 | 0.01 | 0.02 | 0.03 | 0.07 | 0.12 | 0.13 | 0.01 | 0 |
| Vehicle | 0.69 | 0.03 | 0.02 | 0.01 | 0.02 | 0.02 | 0.02 | 0.01 | 0 |
| Vote | 0.93 | 0.01 | 0 | 0 | 0 | 0 | 0.02 | 0.01 | 0 |
| Vowel | 0.97 | 0 | 0.05 | 0.12 | 0.21 | 0.28 | 0.43 | 0.03 | 0.03 |
| Waveform21 | 0.85 | 0.1 | 0.05 | 0.04 | 0.03 | 0.03 | 0 | 0.01 | 0.01 |
| Waveform40 | 0.84 | 0.12 | 0.08 | 0.06 | 0.04 | 0.04 | 0 | 0.02 | 0.01 |
| wholesale | 0.9 | 0.03 | 0.02 | 0.01 | 0 | 0 | 0 | 0 | 0.01 |
| Wine | 0.97 | 0 | 0 | 0.01 | 0.01 | 0.01 | 0.01 | 0 | 0 |
| Sum | | 1.04 | 0.71 | 0.62 | 0.64 | 0.74 | 1.19 | 0.33 | **0.28** |
| Maximum | | 0.18 | 0.12 | 0.12 | 0.21 | 0.28 | 0.43 | 0.05 | **0.04** |

As can be noticed in Table 5, ENN and IINC are the most stable classifiers, while the rest of the algorithms are much less stable, even if they beat (sometimes) ENN and IINC.

## CONCLUSION

This work is a new attempt to enhance the performance of some nearest neighbour classifiers using Hassanat distance metric. The experimental results using a variety of data sets of real life problems have demonstrated the superiority of this distance metric over the traditional and most-used distances, such as Manhattan distance. In addition, we have proved that this distance metric is invariant to data scale, noise and outliers, and therefore, we recommend other researchers use such a distance in other classification problems. Our future work will focus on exploiting and investigating the power of the Hassanat distance metric in other real life problems, such as content-based image retrieval and clustering problems.

## ACKNOWLEDGMENT

All the data sets used in this paper were taken from the UCI Irvine Machine Learning Repository, therefore the authors would like to thank and acknowledge the people behind this great corpus. Also the authors would like to thank the anonymous reviewers of this paper.

## REFERENCES

Alpaydin, E., 1997. Voting Over Multiple Condensed Nearest Neoghbors. *Artificial Intelligence Review*, 11, pp.115-32.

Bache, K. & Lichman, M., 2013. *UCI Machine Learning Repository*. [Online] Available at: http://archive.ics.uci.edu/ml.

Bharkad, S.D. & Kokare, M., 2011. PERFORMANCE EVALUATION OF DISTANCE METRICS: APPLICATION TO FINGERPRINT RECOGNITION. *International Journal of Pattern Recognition and Artificial Intelligence*, 25(6), pp.777-806.

Bhatia, N. & Vandana, A., 2010. Survey of Nearest Neighbor Techniques. *(IJCSIS) International Journal of Computer Science and Information Security*, 8(2), pp.302-05.

Cover, T.M. & Hart, P.E., 1967. Nearest Neighbor Pattern Classification. *IEEE Trans. Inform. Theory*, IT-13, pp.21-27.

Duda, R.O., Hart, P.E. & Stork, D.G., 2001. *Pattern Classification*. 2nd ed. Wiley.

Fix, E. & Hodges, J., 1951. 4 *Discriminatory Analysis: Nonparametric Discrimination: Consistency Properties*. Randolph Field, Texas: USAF School of Aviation Medicine.

Hamamoto, Y., Uchimura, S. & Tomita, S., 1997. A Bootstrap Technique for Nearest Neighbor Classifier Design. *IEEE TRANSACTIONS ON PATTERN ANALYSIS AND MACHINE INTELLIGENCE*, 19(1), pp.73-79.

Hassanat, A.B., 2009. *Visual Words for Automatic Lip-Reading*. PhD Thesis. Buckingham, UK: University of Buckingham.

Hassanat, A.B., 2014. Dimensionality Invariant Similarity Measure. *Journal of American Science*, 10(8), pp.221-26.

Hassanat, A.B., 2014. Solving the Problem of the K Parameter in the KNN Classifier Using an Ensemble Learning Approach. *International Journal of Computer Science and Information Security*, 12(8), pp.33-39.

Jirina, M. & Jirina, M.J., 2008. No. V-1034 *Classifier Based on Inverted Indexes of Neighbors*. Technical Report. Academy of Sciences of the Czech Republic.

Jirina, M. & Jirina, M.J., 2010. Using Singularity Exponent in Distance Based Classifier. In *Proceedings of the 10th International Conference on Intelligent Systems Design and Applications (ISDA2010)*. Cairo, 2010.

Jirina, M. & Jirina, M.J., 2011. Classifiers Based on Inverted Distances. In K. Funatsu, ed. *New Fundamental Technologies in Data Mining*. InTech. Ch. 19. pp.369-87.

Kataria, A. & Singh, M.D., 2013. A Review of Data Classification Using K-Nearest Neighbour Algorithm. *International Journal of Emerging Technology and Advanced Engineering*, 3(6), pp.354-60.

Nathan Krislock, H.W., 2012. *Euclidean Distance Matrices and Applications*. Springer US.

Weinberger, K.Q. & Saul, L.K., 2009. Distance Metric Learning for Large Margin Nearest Neighbor Classification. *Journal of Machine Learning Research*, 10, pp.207-44.

Wu, X.a.V.K.e., 2010. *The top ten algorithms in data mining*. CRC Press.

Yang, L., 2006. *Distance metric learning: A comprehensive survey*. Technical report. Michigan State University.